\documentclass{article}

    \usepackage[final]{ARLET_2024}

\usepackage[utf8]{inputenc} 
\usepackage[T1]{fontenc}    
\usepackage{hyperref}       
\usepackage{url}            
\usepackage{booktabs}       
\usepackage{amsfonts}       
\usepackage{nicefrac}       
\usepackage{microtype}      
\usepackage{xcolor}         
\usepackage{multirow}
\usepackage{subcaption}
\usepackage{graphicx}
\usepackage{array}
\usepackage{float}
\usepackage{amsfonts}
\usepackage{algorithm}
\usepackage[noend]{algpseudocode}
\usepackage{amsmath} 

\makeatletter
\def\algbackskip{\hskip-\ALG@thistlm}
\makeatother

\title{Accelerated Online Reinforcement Learning using Auxiliary Start State Distributions}

\author{%
  Aman Mehra \\
  Robotics Institute\\
  Carnegie Mellon University\\
  Pittsburgh, PA 15213 \\
  \texttt{amanmehr@andrew.cmu.edu} \\
  \And
  Alexandre Capone \\
  Robotics Institute\\
  Carnegie Mellon University\\
  Pittsburgh, PA 15213 \\
  \texttt{acapone2@andrew.cmu.edu} \\
  \AND
  Jeff Schneider \\
  Robotics Institute\\
  Carnegie Mellon University\\
  Pittsburgh, PA 15213 \\
  \texttt{jeff4@andrew.cmu.edu} \\
}

\begin{document}

\newcommand{\am}[1]{{\color{orange} [Aman: #1]}}
\newcommand{\js}[1]{{\color{cyan} [Jeff: #1]}}
\newcommand{\ac}[1]{{\color{green} [Alex: #1]}}

\maketitle

\begin{abstract}
A long-standing problem in online reinforcement learning (RL) is of ensuring sample efficiency, which stems from an inability to explore environments efficiently. Most attempts at efficient exploration tackle this problem in a setting where learning begins from scratch, without prior information available to bootstrap learning. However, such approaches fail to leverage expert demonstrations and simulators that can reset to arbitrary states. These affordances are valuable resources that offer enormous potential to guide exploration and speed up learning. In this paper, we explore how a small number of expert demonstrations and a simulator allowing arbitrary resets can accelerate learning during online RL. We find that training with a suitable choice of an auxiliary start state distribution that may differ from the true start state distribution of the underlying Markov Decision Process can significantly improve sample efficiency. We find that using a notion of safety to inform the choice of this auxiliary distribution significantly accelerates learning. By using episode length information as a way to operationalize this notion, we demonstrate state-of-the-art sample efficiency on a sparse-reward hard-exploration environment. 

\end{abstract}

\section{Introduction}

 Online reinforcement learning algorithms facilitate learning general behaviors without inductive biases and domain expertise through trial and error. By learning from environmental interaction such methods hold the potential to exceed the performance of supervised learning alternatives, reaching superhuman levels of performance on tasks such as Atari \cite{dqn} and Go \cite{alphago}. Despite such successes, these algorithms find it challenging to explore environments efficiently, resulting in long training times \cite{curiosity, go_explore, hyq}. 
 

There has been a considerable amount of work on making online RL more efficient by promoting exploratory behaviors that are novelty-seeking \cite{curiosity} and state space-covering \cite{sac, state_entopy_glen, state_entropy_lili}. Although such approaches have the potential to learn robust policies, the lack of task-directed exploratory cues \cite{barl} and a tendency to forget how to revisit promising exploration frontiers \cite{go_explore} make them inefficient at learning to solve hard-exploration tasks. Moreover, these methods have been designed to improve exploration efficiency in the absence of any other prior information. Consequently, when expert data or a simulator with arbitrary reset conditions are available, these approaches fail to adequately leverage these additional resources to accelerate exploration.

On the other extreme, methods like imitation learning \cite{gail} and offline RL \cite{iql,cql} can learn task-specific behavior from purely offline data. These methods perform well within the distribution of the training data but fail to be robust in an out-of-distribution (OOD) setting, making them unsuitable for use in the real world.

Hybrid RL approaches mix offline data with online interactions to bridge this gap and learn robust policies efficiently. Bootstrapping online training with offline data isn't straightforward and naively finetuning a policy learned offline leads to sub-optimal performance \cite{jsrl}. In particular, offline experience can be quickly forgotten during online training if not handled appropriately \cite{jsrl, hyq}. Successful hybrid methods ensure the persistence of information acquired offline during online training by freezing a part of the replay buffer or learning fixed reference policies using the offline data. 


    
In this paper, we revisit this hybrid RL setup and investigate how limited quantities of expert quality offline data can be used to effectively bootstrap online RL. We show that we can considerably accelerate online learning in a setting where the environment can be reset to arbitrary states by using expert offline data to construct auxiliary start state distributions. 


    
The main contributions of this work are - 
\begin{itemize} 
    \item We show that when an arbitrarily resetable simulator is available we can use a small amount of state information collected from an offline expert to create an auxiliary start state distribution that significantly improves the sample-efficiency of online RL. 
    \item We find that using a notion of safety, approximated via episode length information is crucial for forming auxiliary start state distributions that accelerate training. Moreover, we  show that this yields policies more robust to shifts in the start state distribution.
    \item We present empirical results on a hard-exploration sparse-reward maze environment and show state-of-the-art sample efficiency and robustness.
\end{itemize}

\section{Related Work}

We explore related literature in this space through three broad category of methods: i) purely online RL, ii) purely offline learning and iii) hybrid RL methods.

\textbf{Exploration in purely online RL:} Exploration is an age-old problem in reinforcement learning that has received significant attention in the online RL context. Several methods inject additive noise to the actions \cite{ppo} or network parameters \cite{rnd} to perform exploration. Such exploration is incidental to the primary objective of reward maximization and not very efficient at exploring the state space \cite{hyq}. Many approaches incentivize exploratory behaviour through exploration bonuses such as surprise-maximizing intrinsic motivation \cite{curiosity}, surprise-minimizing intrinsic motivation \cite{surp_minimize}, and action \cite{sac}, state \cite{state_entropy_lili} and trajectory \cite{traj_entropy_explore} entropy maximizing rewards.  Entropy maximization approaches fail to distinguish exploration in unseen regions from exploration in regions of the state space where the policy is already proficient. This makes them inefficient. While intrinsic motivation based methods are guided by a notion of surprise, they too struggle in hard-exploration sparse-reward environments \cite{go_explore}. Moreover, these methods are unable to leverage affordances like offline data and resetable simulators when available.

Go-explore \cite{go_explore} is a conceptual framework that disentangles the question of where to explore from how to get there. It is reminiscent of classical planners that first choose an exploration frontier, navigate to it quickly without exploring (or by resetting the simulator to that frontier state) and then initiate exploration after arriving at the frontier. Go-explore maintains an archive of visited states and chooses an exploration frontier from this archive either uniformly at random or using a domain specific heuristic. Our work follows a similar theme but extends this framework by investigating what is a good way of picking an exploration frontier. We present generic properties that are desirable to have in this selection procedure and present a mechanism to select exploration frontiers that will be broadly applicable across a range of tasks. 

BARL \cite{barl} is an information theoretic exploration method that uses a classical planner and a learnt posterior model to sample transitions that are maximally informative for the policy to learn a given task. This enables it to solve tasks very efficiently. The setting used by its authors bares close resembles to ours since they assume access to a simulator that supports arbitrary resets. Moreover, their use of a Gaussian process (GP) to model the posterior is amenable to utilizing expert demonstrations during training. While very effective on small scale problems with dense reward functions, BARL unfortunately does not scale to higher dimensions and sparse-reward settings. This is confirmed by us in our experiments.

\textbf{Learning policies efficiently offline:} Another way to efficiently learn policies is by training on a purely offline dataset of experiences. This sidesteps the issue of online exploration and efficiently recovers a policy based on the offline dataset. Methods such as behaviour cloning and GAIL \cite{gail} fall under the broad class of imitation learning algorithms that model policy learning as a supervised learning problem and learn a mapping from states to actions. A key issue with imitation learning methods is that they are highly brittle and require access to large amounts of high quality expert data to succeed \cite{bridge_offline_online}. 

Offline reinforcement learning \cite{offlineRLtut} is another offline learning paradigm capable of efficiently learning policies from demonstration data of mixed quality while requiring good state coverage in the offline dataset \cite{offlineRLvsBC, bridge_offline_online}. Consequently, a key challenge with this approach is that the lack of online interactions leaves offline RL susceptible to distribution shift. Wrongly estimating values for actions beyond the support of the  dataset can hamper training \cite{offlineRLtut}. Conservative Q-learning (CQL) \cite{cql} is a recent offline RL method that attempts to tackle this problem by maintaining pessimism within the Q-value function towards actions that are absent from the offline dataset. Implicit Q-learning (IQL) \cite{iql} completely avoids predicting value estimates for unseen actions by learning a distributional state-value function and computing an upper expectile over it to obtain the value estimate of the best action in that state. Real world deployment of these algorithms invariably lead to encounters with OOD states and actions making online finetuning a necessity for the real world deployment of such algorithms.

\textbf{Hybrid Reinforcement Learning: } Hybrid reinforcement learning leverages a combination of offline data with online interaction to learn policies. The main challenge in hybrid reinforcement learning is to devise methods that effectively bootstrap online learning from offline data. Several approaches \cite{Rajeswaran2017LearningCD, dqfd} do this by using imitation to learn a policy from offline demonstrations before finetuning it with RL. However, most modern state-of-the-art online RL algorithms are value based \cite{sac, ppo}. Naively finetuning an offline acquired policy with value based RL can cause significant performance degradation as a value function of similar quality to the pretrained policy is not available at the start of online finetuning \cite{jsrl}. Though Monte Carlo return estimate based algorithms exist, their online finetuning is known to be less efficient \cite{awac}. Offline RL presents a transferable paradigm to train a policy and value function with identical objectives in both offline and online setups. However, not all offline RL methods are well suited for online finetuning due to their inherent pessimism towards distribution shift \cite{awac}. Even better suited offline RL methods like IQL \cite{iql} result in weaker policies after finetuning especially when limited offline data is available \cite{jsrl}. An alternative line of work \cite{hyq, nair2018overcoming, vecerik2017leveraging} avoids finetuning a pretrained policy all together by pre-filling replay buffers at the start of training with transitions from the offline dataset. These transitions persist through training and policy learning happens from scratch. JSRL \cite{jsrl} presents an alternative approach to the finetuning-free idea of hybrid RL. It learns a guide policy from offline data and uses it to roll out a part of the online episode before handing over control to a freshly initialized policy for completing the roll-out. The handover point is altered over the course of training and all the captured experience is used to train the freshly initialized policy. 

The proposed work also lies in this finetuning-free hybrid setting and shares similarities with JSRL. Both the proposed work and JSRL conceptually belong to the Go-explore \cite{go_explore} family of algorithms. The two key differences between JSRL and the proposed work are: i) both works encapsulate the idea of a reset distribution or equivalently a frontier state to explore from. While ours is reached through environmental resets, JSRL uses a guide policy to reach it. ii) JSRL induces a specific kind of reset distribution through its variation of the handover point over the course of training. This differs from our reset distribution and we will show in our method and experiments that this is an important choice that influences the sample-efficiency and performance of the learnt policy.



\section{Preliminaries}

We define a finite-horizon discrete-time MDP $\mathcal{M}$ to be a ($\mathcal{S},\mathcal{A},r, p_0 ,H,\mathcal{T},\gamma$) tuple where  $\mathcal{S}$ is the state space, $\mathcal{A}$ is the action space, $r: \mathcal{S} \times \mathcal{A} \times \mathcal{S} \rightarrow \mathbb{R}$ is the reward function, $p_0$ is a probability distribution defined over $\mathcal{S}$ corresponding to the start state distribution of $\mathcal{M}$, $H \in \mathbb{N}$ is the finite time horizon, $\gamma$ is the discount factor and $\mathcal{T} : \mathcal{S} \times \mathcal{A} \rightarrow P(\mathcal{S})$ describes a transition function capturing the distribution of next states when an action $a \in \mathcal{A}$ is taken at a state $s \in \mathcal{S}$. 

In RL, the goal is to obtain a policy $\pi(a|s)$ that maximizes the expected sum of future discounted rewards from $p_0$. Concretely, the objective is to maximize -
\begin{equation}
    \label{eqn:reward}
    \mathcal{J}_{p_0}(\pi) = \mathbb{E}_{s_0 \sim p_0, s_{t+1} \sim \mathcal{T}(s_t,a_t), a_t \sim \pi(s_t)}[ \Sigma_{t=0}^{H} \gamma^t r(s_t,a_t, s_{t+1}) ]
\end{equation}

It is important to note here that, while $\mathcal{J}_{p_0}({\pi})$ is maximized with respect to $p_0$, it is not necessary to train with this start state distribution. As discussed in \cite{kakade2002}, $p_0$ can down-weight the influence of unlikely but important states during policy improvement by visiting them infrequently. The presence of such unlikely but important states along the optimal trajectory, which we refer to as \emph{task critical states} ($\mathcal{C}$), can result in slow learning progress. It would instead be beneficial to have a start state distribution ($\mu$) that is distributed more uniformly across some subset $\mathcal{S}_{sub}$ of the state space (i.e., $\mathcal{S}_{sub} \subseteq \mathcal{S}$) emphasizing visitation of \emph{task critical states} while asymptotically obtaining comparable performance to the optimal policy as measured by $\mathcal{J}_{p_0}(\pi)$. 

In addition, it can be noted that robustness is another desirable property for a learnt policy to have. More precisely, we would like a policy learnt to maximize $\mathcal{J}_{p_0}$ to also do well on a broader start state distribution which we call $\mu_{OOD}$, i.e., maximize $\mathcal{J}_{\mu_{OOD}}$ (\textit{OOD} refers to Out-Of-Distribution). Thus evaluating robustness signifies evaluating the learnt policy on a separate start state distribution defined by $\mu_{OOD}$. The state distribution induced by a policy starting from $\mu_{OOD}$ may be different from the induced distribution of states when starting from $p_0$. As a result proficiency starting from $p_0$ does not imply proficiency starting from $\mu_{OOD}$. For policies designed for real world deployment, it is important for them to generalize beyond the training distribution making evaluation from $\mu_{OOD}$ a useful benchmark to consider while training. 

In this paper, we investigate how sample-efficiency of online RL can be enhanced by suitably choosing $\mu$ and how this impacts robustness. Unlike traditional RL where policy rollouts are constrained to begin from $p_0$, we assume access to a simulator that allows resets to any arbitrary start state distribution $\mu$. This is a common affordance available in a wide range of RL simulators that is not fully utilized by existing algorithms. As we will show in Section~\ref{sec:auxss}, the reset states belong to a subset $\mathcal{S}_{sub}$ of the state space that spans the offline expert demonstration distribution ($\mu_{off}$) only. 

It is important to note that since $\mu_{off}$ is generated via an expert starting from the distribution $p_0$, the state distribution of $\mu_{off}$ will be similar to the induced state distribution of a well performing policy starting from $p_0$. As a result, resetting the simulator to a $\mu$ derived from $\mu_{off}$ will not compromise the robustness benchmark. At the start of training, novel states emerging from changing reset distributions from $p_0$ to $\mu_{OOD}$ will remain novel when changing reset distributions from $\mu$ to $\mu_{OOD}$. Over the course of training, resetting to $\mu$ may simply alter the likelihood of visiting some novel states. The impact of this altered likelihood will be observable through the robustness evaluations.

\section{Auxiliary Start States for Accelerated Learning}
\label{sec:auxss}

Directly computing the visitation distribution over \emph{task critical states} ($\mathcal{C}$) is challenging in continuous state spaces owing to the associated computational infeasibility of calculating this quantity. Moreover, this is a policy dependant quantity and as a result will need to be continuously recomputed over the course of training. As a result, a suitable $\mu$ should be an easily computable dynamic distribution that accounts for the changing visitation distribution of the policy over time.

We observe that episode termination is a powerful and ubiquitous signal available in a variety of RL tasks. It is especially prevalent in robotic tasks such as autonomous driving and robot locomotion where eventual real world deployment is the end goal and safety of the agent and its surroundings are of paramount importance.

Intuitively, for a state $s$, if the proportion of actions that cause the agent to land in a terminal state is high, then a larger exploratory budget is required to learn a feasible action for this state by the policy. Moreover, the chance of navigating through this state likely hinges on the repeated selection of a small set of safe and feasible actions in the neighbouring regions of the state space. Therefore, there is a high likelihood that such states belong to $\mathcal{C}$ and sampling them more frequently can accelerate learning. More formally, we define the notion of state safety for any state $ s \in \mathcal{S}$ as - 

\begin{equation}
    \Omega_{\pi}(s) = \int_{a_{0:k-1}} P(a_{0:k-1} | s, \pi) \int_{s_k} P(s_k | s, a_{0:k-1}, \mathcal{T}, \pi) \mathcal{Z}(s_k) \,ds_k  \,da_{0:k-1}
\end{equation}

Here, $\mathcal{Z}(s) \in \{0,1\} \, \forall s \in \mathcal{S}$ and denotes whether or not state $s$ causes episode termination.   $\mathcal{Z}(s) = 0$ if episode termination is caused by being in state $s$ and $1$ otherwise. $a_{0:k-1}$ is the sequence of $k$ actions induced by the policy $\pi$ from state $s$ under the transition model given by $\mathcal{T}$. $s_k$ is the state that is reached when policy $\pi$ takes action sequence $a_{0:k-1}$ starting from state $s$ in an environment with transition model $\mathcal{T}$.

Exactly computing $\Omega_{\pi}(s)$ is still computationally expensive. Instead we leverage the time to termination or episode length from a given start state, which is a freely available metric at training time, as a Monte Carlo approximation of the true state safety for a given policy at that state. By maintaining a parameterized distribution over a set of desirable start states (\emph{candidate task critical states} or $\mathcal{\widetilde{C}}$) we can exploit local smoothness in the majority of the state space to quickly propagate these approximations across the start state distribution (see Algorithm~\ref{algo:auxSS_update} for details)

Since we are taking a hybrid approach to RL, we have access to a limited amount of expert demonstration data. Since this data comprises successful demonstrations of the task, the demonstration trajectories will likely contain \emph{task critical states} if they exist. As a result, we can simply set $\mathcal{\widetilde{C}}$ to be the states from the demonstration data and identify $\mathcal{C}$ from amongst these states over the course of training. Putting everything together we get our proposed method \textit{AuxSS} that we describe in Algorithm~\ref{algo:auxSS}.

\begin{algorithm}
\caption{Online RL with Auxiliary Start States}
\label{algo:auxSS}
\hspace*{\algorithmicindent} 
\hspace*{\algorithmicindent}  
\begin{algorithmic}[1]

\State \textbf{Inputs:} Task Horizon $H$, Offline Demonstration States $\mathcal{S}_{demo}$, Algorithm $\mathcal{A}$, Training Timesteps $T_{max}$, Environment $\mathcal{E}$, replay buffer $\mathcal{B}$ 

\State Sampling distribution $\mathcal{W} \gets \overbrace{ [1, 1 \, ... \, 1] }^\text{$\mathrm{len}(\mathcal{S}_{demo})$}$ \Comment{Initialization incentivizes visiting states atleast once}
\State Sampling distribution norm $\mathcal{N} \gets \Call{sum}{\mathcal{W}}$
\State $t \gets 0$
\While{$t \leq T_{max}$}
\State $i \gets \Call{SampleStartState}{\frac{\mathcal{W}}{\mathcal{N}}}$
\State $s_0 \gets \mathcal{S}_{demo}[i]$
\State $L_{ep} \gets \Call{TrainForOneEpisode}{\mathcal{A}, \mathcal{B}, \mathcal{E}, s_0)}$ \Comment{Return value is episode length}
\State $t \gets t + L_{ep}$
\State $\mathcal{W}, \mathcal{N} \gets \Call{UpdateSampler}{\mathcal{W}, \mathcal{N}, L_{ep}, i, H, \mathcal{S}_{demo}}$ \Comment{See Algorithm~\ref{algo:auxSS_update}}
\EndWhile
\end{algorithmic}
\end{algorithm}

\begin{algorithm}
\caption{Updating Auxiliary Start State Distribution via Episode Length (\textit{AuxSS})}
\label{algo:auxSS_update}
\hspace*{\algorithmicindent} 
\hspace*{\algorithmicindent}  
\begin{algorithmic}[1]

\State \textbf{Inputs:} Sampling distribution $\mathcal{W}$, Sampling distribution norm $\mathcal{N}$, Episode Length $L_{ep}$, Update index $i$, Task Horizon $H$, Offline Demonstration States $\mathcal{S}_{demo}$, Weight Threshold $\delta$, Smoothing Variance  $\sigma^2$

\State \textbf{Outputs:} Sampling distribution $\mathcal{W}$, Sampling distribution norm $\mathcal{N}$ 

\State  $\mathcal{W}[i] \gets \Call{Max}{\frac{H - L_{ep}}{H}, \delta}$ \Comment{$\delta$ ensures probability of sampling $ \geq 0$}

\State $\lambda \gets \frac{1}{\sqrt{2 \pi} \sigma} \Call{exp}{\frac{(S_{demo} - S_{demo}[i])^2}{2 \sigma^2}} $ \Comment{$\lambda$ is used for smoothing updates to $\mathcal{W}$ }

\State  $\mathcal{W} \gets (1 - \lambda)\mathcal{W}  + \lambda \mathcal{W}[i]  $

\State $\mathcal{N} \gets \Call{sum}{\mathcal{W}}$

\end{algorithmic}
\end{algorithm}

\section{Experiments}

\textbf{Overview:} In this section we first highlight the affinity of using auxiliary start state distributions with the hybrid RL setup by demonstrating state-of-the-art sample-efficiency on a sparse-reward hard-exploration task. We show that \textit{AuxSS} is better suited at assimilating information from limited amounts of expert offline data by demonstrating better sample-efficiency than competing approaches that have access to $15\times$ more offline expert data available to them. We simultaneously show \textit{AuxSS} also facilitates learning more robust policies. Finally, we empirically demonstrate that approximating a more uniform visitation distribution over $\mathcal{C}$ through $\Omega$ facilitates accelerated learning. We showcase how \textit{AuxSS} is a good way to approximate $\Omega$ while other distributions not motivated in the same way are not.



\textbf{Setup:} We conduct our experiments on the maze setup shown in Figure ~\ref{fig:ss_dists}. It consists of two large regions connected by a narrow traversable passage with obstacles on either side (shown in blue). The obstacles are pits of of lava and the episode terminates if an agent encroaches this area. We refer to this environment as Lava Bridge. It has a continuous $4D$ state space comprising $2$ dimensions of position and $2$ dimensions of velocity. The action space is a continuous $2D$ force vector. All evaluations are performed in a sparse-reward setup where the agent only gets a non-zero reward on reaching the goal state or entering a terminal state.

\begin{figure*}
  \centering
    \includegraphics[width=\textwidth]{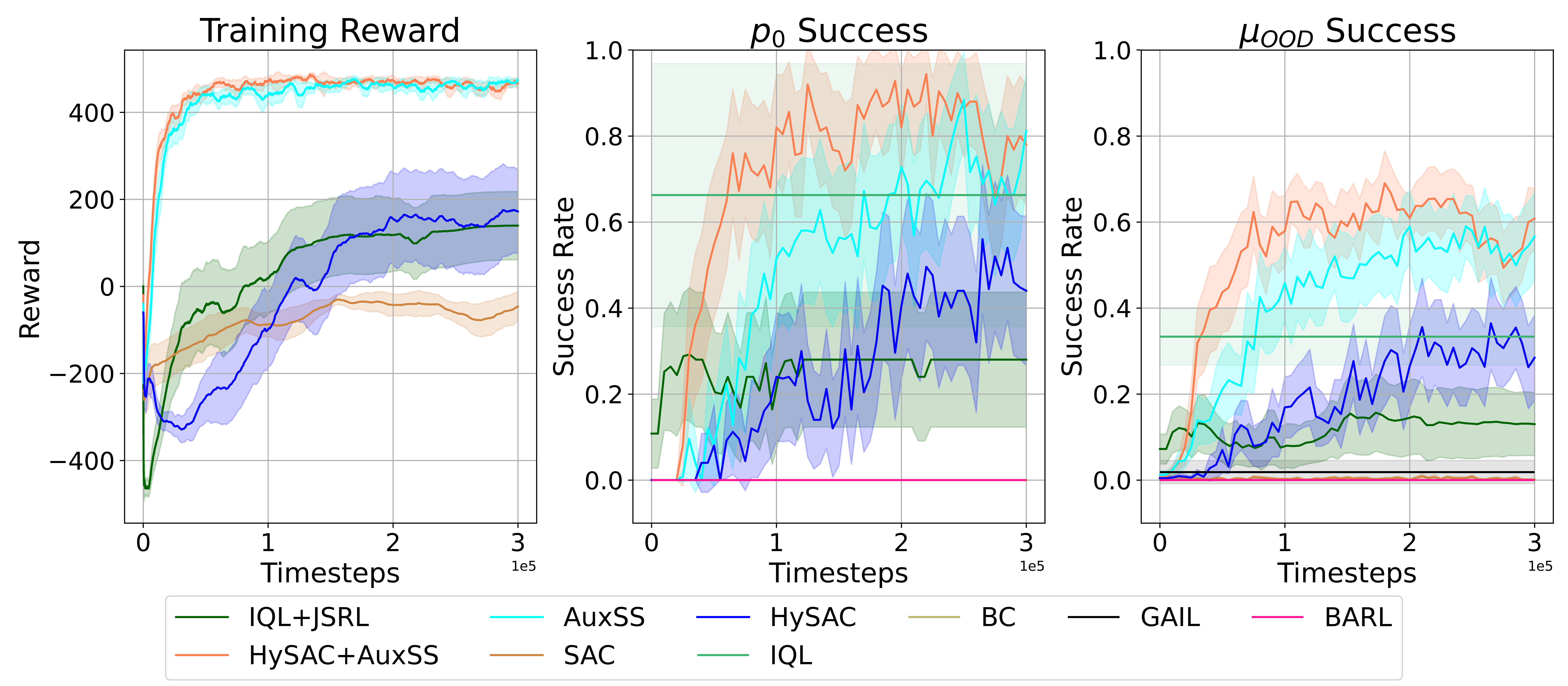}
    \caption{Task Completion Rate of Various methods on the Lava Bridge Environment. Each method is evaluated on an In Distribution (ID) and Out-of-Distribution (OOD) benchmark of starting states where the ID start state distribution is the start state distribution of the MDP while the OOD benchmark comprises a different distribution of start states.}
    
  \label{fig:lava_bridge_benchmark}

\end{figure*}

\subsection{Do Auxiliary Start State Distributions accelerate learning of robust policies?}


In this section we study the efficacy of \textit{AuxSS} at improving the sample efficiency of online learning by making use of affordances such as arbitrary resetting of the environment and access to a limited quantity of expert demonstration data. We compare \textit{AuxSS} with a variety of offline, online and hybrid methods. In addition to tracking sample-efficiency we also track the robustness of the learnt policy. The choice of the MDP start state distribution ($p_0$) and robustness benchmark start state distribution ($\mu_{OOD}$), are shown in Figure~\ref{fig:ss_dists}.

Figure~\ref{fig:lava_bridge_benchmark} presents the findings of this study on the Lava Bridge environment. A standardized training setup has been used across methods (except BARL \cite{barl}) where the number of offline demonstration transitions is set to $10$ million, number of online learning steps is $300000$, replay buffer size is $10000$, max episode length is $500$ and experiments are evaluated across $25$ seeds. All hybrid methods and offline methods have access to $500$ transitions of expert demonstration data.

We use the example of SAC \cite{sac}, a purely online RL method, to highlight the exploration challenges that the Lava Bridge environment poses to standard online RL. SAC uses an undirected entropy based bonuses to promote exploration but struggles to efficiently explore in our environment. Its failure to reach the goal within the stipulated training budget and learn a robust policy highlights the exploration challenges posed by the Lava Bridge environment. In addition we evaluate BARL \cite{barl}, an information theoretic method for sample-efficient online exploration. We evaluated BARL both with and without access to the demonstration data and found it unable to solve our task. BARL is reliant on a classical planner which is designed to work with dense rewards. Therfore the sparse-reward nature of our task prevents the BARL planner from finding solutions in a feasible amount of time, resulting in the method's failure.

We compare our proposed approach with two hybrid RL approaches - HySAC (an adaptation of HyQ \cite{hyq} where a DQN is replaced with SAC) and JSRL \cite{jsrl}. It can be seen that across training reward and success rates, using a good auxiliary start state distribution yields state-of-the-art sample efficiency and performance as both HySAC and JSRL struggle to make full use of the limited offline demonstration data. Moreover, the proposed approach is complementary to HySAC's persistent storage of offline demonstration in the replay buffer throughout training and as a result the two approaches can be coupled (HySAC+\textit{AuxSS}) to obtain better robustness in fewer training steps. 


\begin{figure*}
  \centering
    \includegraphics[width=\textwidth]{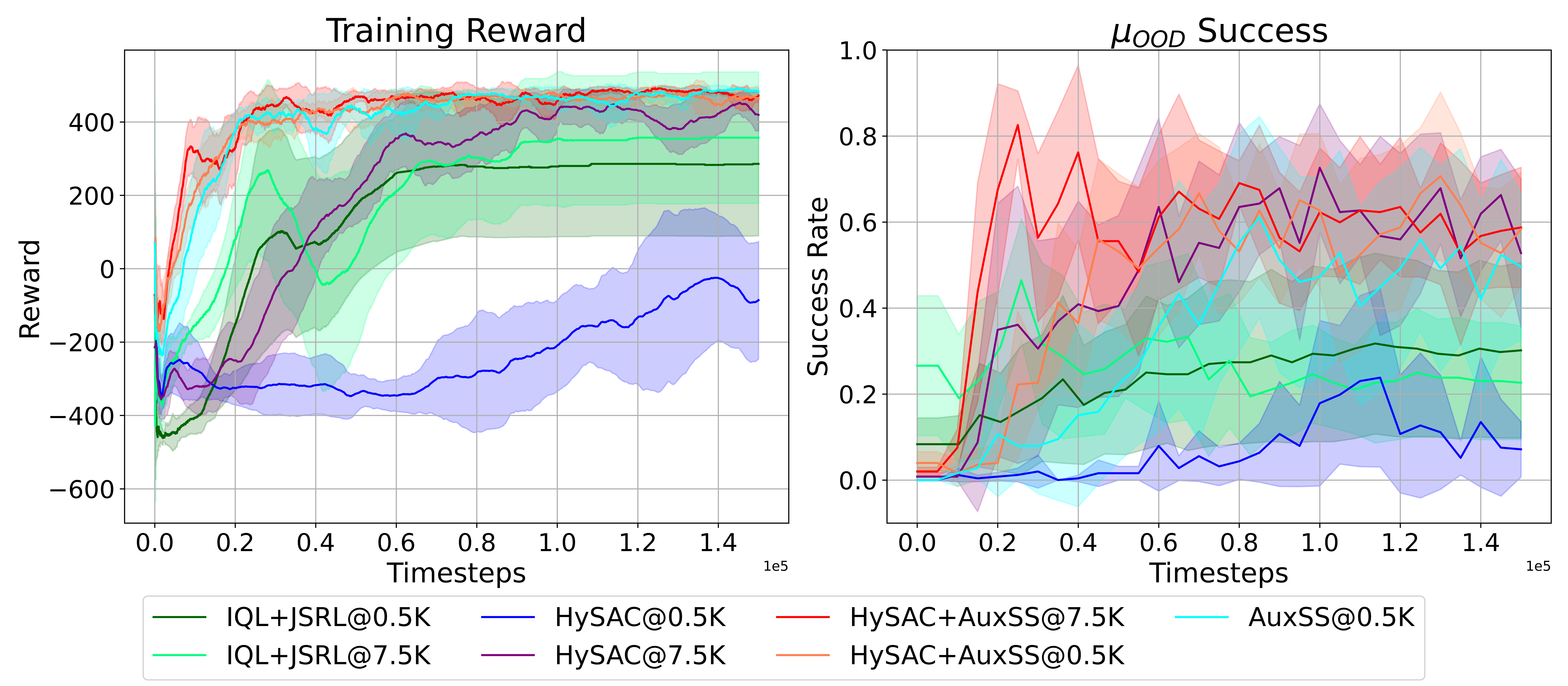}
    \caption{A study of how sample-efficiency and robustness vary for hybrid RL methods when provided with different amounts of demonstration data.}
  \label{fig:lava_bridge_demo_size}
\end{figure*}

It can be noted that the approach taken by JSRL of handing over episode rollout from a guide policy to the learning policy is conceptually similar to having an auxiliary start state distribution that monotonically recedes towards $p_0$ over the course of training. Unlike our proposed auxiliary distribution, JSRL cannot reemphasize visiting previously learnt regions of the state space that may have been forgotten over the course of training. Even accounting for some sample-efficiency gains that JSRL may obtain by directly resetting to the handover point (rather than using the guide policy to get there) withing an episode, its inability to reemphasize visitation of previously learnt regions prevent it from learning very robust policies as can be seen in Figure~\ref{fig:lava_bridge_benchmark}.

We also compare against purely offline methods such as imitation learning and offline RL. We find that imitation learning methods like behaviour cloning (BC) and GAIL \cite{gail} fail to succeed on both $p_0$ and $\mu_{OOD}$. This stems from their inability to learn meaningful policies from just $500$ transitions. Infact, we provide these methods with $7.5$K transitions to learn from in Figure~\ref{fig:lava_bridge_benchmark} and they still fail to learn a reasonable policy as measured by both start state distributions. Offline RL fairs a lot better. IQL \cite{iql} is a popular offline RL method that we compare against. We find that IQL performs well on $p_0$ but suffers a large drop in performance when evaluated on $\mu_{OOD}$. This arises from the fact that offline RL methods are incentivized to learn a good policy only within the distribution of their training data which in this case is $\mu_{off}$. For a policy that performs well starting from $p_0$, the induced state distribution of this policy will be close to $\mu_{off}$ and as a result a policy learnt using IQL does well on $p_0$. On the other hand, the induced state distribution of the policy learnt via IQL when starting from $\mu_{OOD}$ would present a distribution shift with respect to $\mu_{off}$ resulting the in the poor robust performance of IQL.

\subsection{Influence of offline demonstration set size on performance and sample-efficiency}

In Figure~\ref{fig:lava_bridge_demo_size} we plot the training reward and evaluate robustness when different quantities of expert demonstration data ($0.5K$ and $7.5K$ expert samples) are available prior to the online learning phase. We find that by accessing $15 \times$ fewer expert samples \textit{AuxSS} and HySAC+\textit{AuxSS} can match and exceed the robustness and sample efficiency of policies learnt via other hybrid RL methods. When provided access to a resetable simulator, this demonstrates that a good auxiliary start state distribution can more effectively assimilate data to guide exploration and accelerate learning than other approaches to hybrid RL. Unlike other methods, having a good start state distribution prevents the need to collect large quantities of expert data through ability to bootstrap online learning off of very limited demonstration trajectories (the $500$ expert transitions come from $3$ demonstration trajectories. Reported \textit{AuxSS} trends sample from a random subset of $150$ transitions. This is approximately one trajectory worth of expert data).


\subsection{State safety inspired start state sampling for sample efficiency}


\begin{figure*}
  \centering
    \begin{subfigure}[t]{0.69\textwidth}
        \centering
        \includegraphics[width=\textwidth]{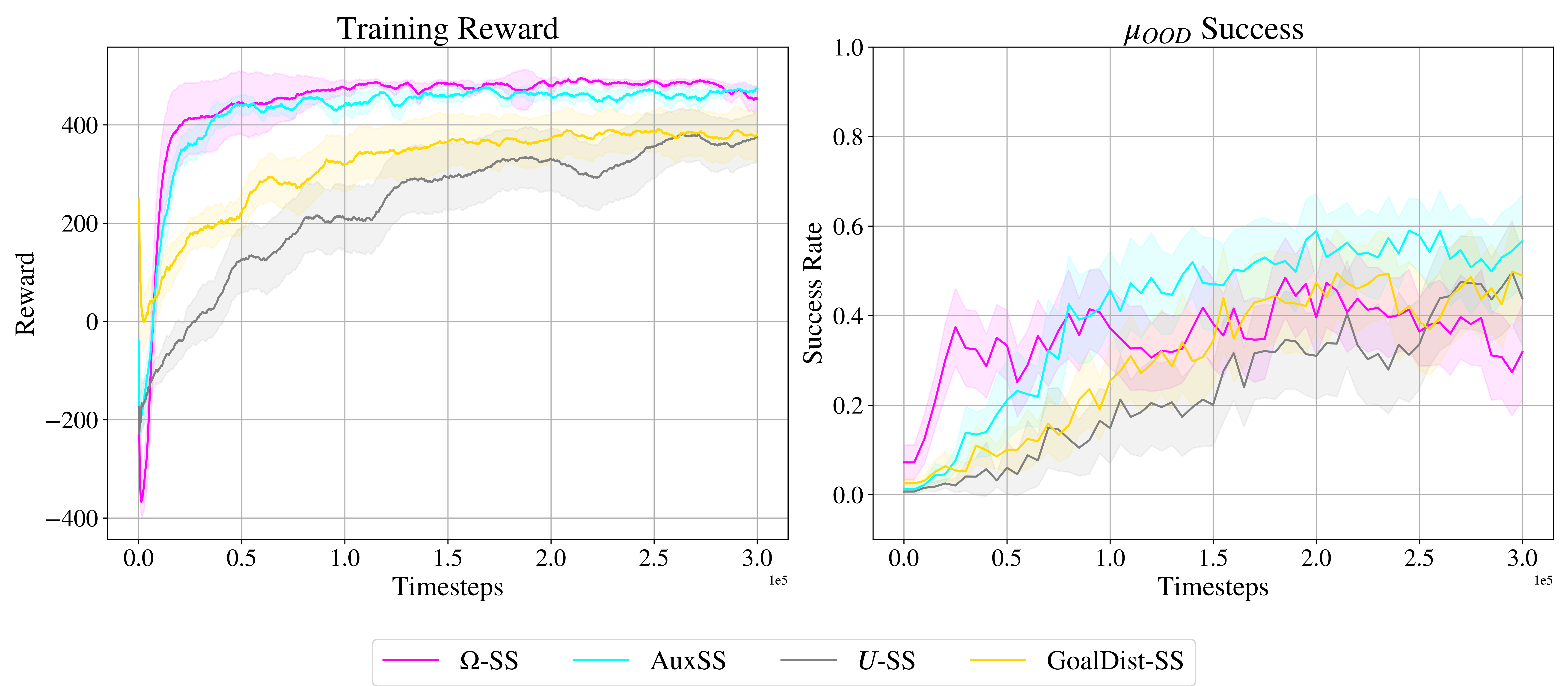}
    \end{subfigure}%
    ~ 
    \begin{subfigure}[t]{0.31\textwidth}
        \centering
        \includegraphics[width=\textwidth,trim={1cm -1.5cm 1cm 0.26cm},clip]{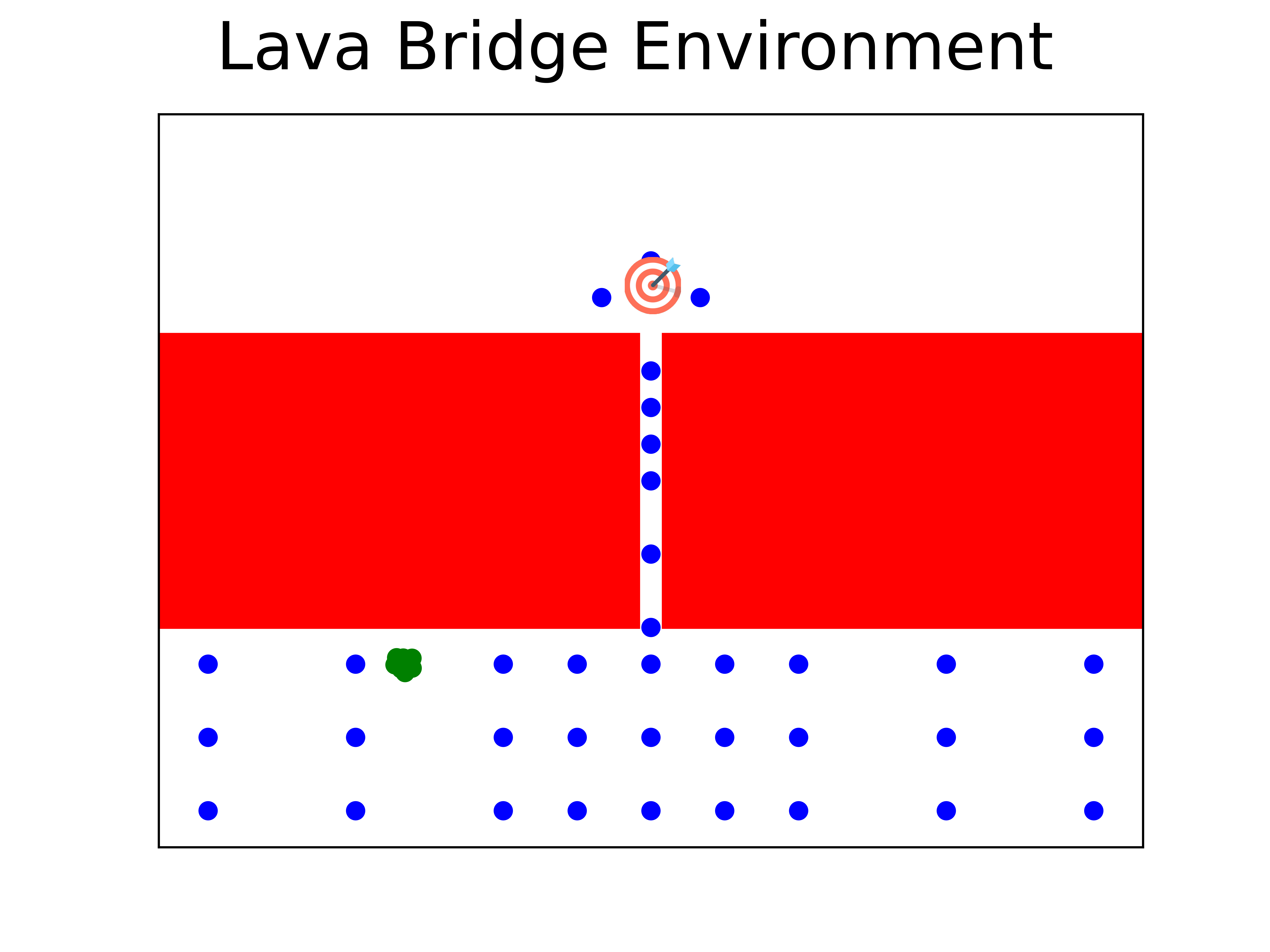}
    \end{subfigure}
    \caption{(\textit{Left}) Sample-efficiency and robustness trends when simulator resets are selected using different start state distributions. (\textit{Right}) An illustration of the Lava Bridge environment. The red regions are the lava pits, the green blobs denote $p_0$ and the blue spots correspond to the distribution $\mu_{OOD}$. The red target marks the goal location.}
    \label{fig:ss_dists}
\end{figure*}

In Section~\ref{sec:auxss}, we connect the notion of state safety $\Omega$ with \textit{task critical states} ($\mathcal{C}$) and discuss how this can influence sample-efficiency. In this section, we empirically validate our claims. We modify \textit{AuxSS} by constructing a static distribution ($\Omega$-SS) that sample start states with respect to a random policy. Concretely, we sample start states inversely proportional to $\Omega_{\pi_{rand}}(s)$ where $\pi_{rand}(.|s) \sim \mathcal{U^{|A|}}$. In practice, we use Monte Carlo sampling of actions for a fixed time horizon ($=4$ time steps) to approximate this quantity for each state. Since the policy at the start of online training is initialized randomly, this mimics the state safety distribution with respect to the policy at the start of training. Therefore if our claims hold we expect to see matching sample-efficiency trends to \textit{AuxSS} in the early stages of training.


Figure~\ref{fig:ss_dists} presents the findings of this study. We see that as expected, $\Omega$-SS demonstrates matching sample-efficiency and robustness trends as \textit{AuxSS} early in training. In fact, since $\Omega$-SS is the correct state safety distribution with respect to the initialized policy from the start of training it learns even faster than \textit{AuxSS}, since \textit{AuxSS} must gradually approximate this state safety distribution over the course of multiple training episodes. 

We note here the divergence in robustness trends seen later in training. This is caused by the static nature of $\Omega$-SS which fails to adapt to the morphing $\mathcal{C}$ induced by the policy as it trains. This causes the resulting loss of robustness. The dynamic nature of \textit{AuxSS} helps prevent this degradation as its able to adapt its start state distribution based on changes in the policy.

\subsection{Do start state distributions not deriving from state safety fail to be sample efficient?}

To study this inverse logical question, we construct two start state distributions, $\mathcal{U}$-SS and \textit{GoalDist}-SS, that do not try to incentivize visitation of \textit{task critical states}. $\mathcal{U}$-SS is a static distribution that uniformly samples states from the provided demonstrations. \textit{GoalDist}-SS is a dynamic distribution that exponentially weights states based on their distance from the task goal. States closer to the goal are assigned a higher probability to be sampled. The time varying component of this distribution arises from temperature scaling of the distribution with the temperature gradually rising over the course of training. This promotes sampling near goal states early on in training and sampling more uniformly from the demonstration data later on in training.

Figure~\ref{fig:ss_dists} contains the findings of this study. It can be seen that both $\mathcal{U}$-SS and \textit{GoalDist}-SS are far slower to train than state safety inspired distribution demonstrating that not all start state distributions will accelerate learning. As a consequence of the poor choice of their state visitation, these methods fail to learn good policies in the stipulated training budget and thus also have much lower robust performance than \textit{AuxSS} and $\Omega$-SS (before its static nature causes robustness to degrade).

\section{Discussion and Limitations}

In this work, we explore the use of commonly available affordances in RL tasks to guide online exploration. We highlight the importance of auxiliary start state distributions, constructed by utilizing small quantities of expert demonstration comprising only state information, in facilitating sample-efficient learning of robust policies. We find that in environments that allow arbitrary state resetting, this is a very crucial design choice and we observe that deriving start state distributions from notions of state safety can dramatically accelerate policy learning online. Naturally, the need for a simulator that supports arbitrary state resets can also be viewed as a limitation. An interesting future direction would be to explore approaches like offline RL to relax this requirement by providing a means to efficiently navigating the distribution of offline data starting from the start state distribution of the MDP.



\bibliography{icmlw_2024}

\begin{thebibliography}{10}

\bibitem{surp_minimize}
G.~Berseth, D.~Geng, C.~M. Devin, N.~Rhinehart, C.~Finn, D.~Jayaraman, and S.~Levine.
\newblock {\{}SM{\}}irl: Surprise minimizing reinforcement learning in unstable environments.
\newblock In {\em International Conference on Learning Representations}, 2021.

\bibitem{rnd}
Y.~Burda, H.~Edwards, A.~Storkey, and O.~Klimov.
\newblock Exploration by random network distillation.
\newblock In {\em International Conference on Learning Representations}, 2019.

\bibitem{go_explore}
A.~Ecoffet, J.~Huizinga, J.~Lehman, K.~O. Stanley, and J.~Clune.
\newblock First return, then explore.
\newblock {\em Nature}, 2021.

\bibitem{sac}
T.~Haarnoja, A.~Zhou, P.~Abbeel, and S.~Levine.
\newblock Soft actor-critic: Off-policy maximum entropy deep reinforcement learning with a stochastic actor.
\newblock In {\em Proceedings of the 35th International Conference on Machine Learning}, 2018.

\bibitem{dqfd}
T.~Hester, M.~Vecerik, O.~Pietquin, M.~Lanctot, T.~Schaul, B.~Piot, D.~Horgan, J.~Quan, A.~Sendonaris, I.~Osband, G.~Dulac-Arnold, J.~Agapiou, J.~Leibo, and A.~Gruslys.
\newblock Deep q-learning from demonstrations.
\newblock {\em Proceedings of the AAAI Conference on Artificial Intelligence}, 2018.

\bibitem{gail}
J.~Ho and S.~Ermon.
\newblock Generative adversarial imitation learning.
\newblock In {\em Advances in Neural Information Processing Systems}, 2016.

\bibitem{state_entopy_glen}
A.~K. Jain, L.~Lehnert, I.~Rish, and G.~Berseth.
\newblock Maximum state entropy exploration using predecessor and successor representations.
\newblock In {\em Advances in Neural Information Processing Systems}, 2023.

\bibitem{traj_entropy_explore}
A.~K. Jain, L.~Lehnert, I.~Rish, and G.~Berseth.
\newblock Maximum state entropy exploration using predecessor and successor representations.
\newblock In {\em Thirty-seventh Conference on Neural Information Processing Systems}, 2023.

\bibitem{kakade2002}
S.~Kakade and J.~Langford.
\newblock Approximately optimal approximate reinforcement learning.
\newblock In {\em International Conference on Machine Learning}, 2002.

\bibitem{iql}
I.~Kostrikov, A.~Nair, and S.~Levine.
\newblock Offline reinforcement learning with implicit q-learning.
\newblock In {\em International Conference on Learning Representations}, 2022.

\bibitem{offlineRLvsBC}
A.~Kumar, J.~Hong, A.~Singh, and S.~Levine.
\newblock Should i run offline reinforcement learning or behavioral cloning?
\newblock In {\em International Conference on Learning Representations}, 2022.

\bibitem{cql}
A.~Kumar, A.~Zhou, G.~Tucker, and S.~Levine.
\newblock Conservative q-learning for offline reinforcement learning.
\newblock In {\em Advances in Neural Information Processing Systems}, 2020.

\bibitem{offlineRLtut}
S.~Levine, A.~Kumar, G.~Tucker, and J.~Fu.
\newblock Offline reinforcement learning: Tutorial, review, and perspectives on open problems.
\newblock {\em arXiv preprint arXiv:2005.01643}, 2020.

\bibitem{barl}
V.~Mehta, B.~Paria, J.~Schneider, W.~Neiswanger, and S.~Ermon.
\newblock An experimental design perspective on model-based reinforcement learning.
\newblock In {\em International Conference on Learning Representations}, 2022.

\bibitem{dqn}
V.~Mnih, K.~Kavukcuoglu, D.~Silver, A.~A. Rusu, J.~Veness, M.~G. Bellemare, A.~Graves, M.~Riedmiller, A.~K. Fidjeland, G.~Ostrovski, et~al.
\newblock Human-level control through deep reinforcement learning.
\newblock {\em nature}, 2015.

\bibitem{awac}
A.~Nair, A.~Gupta, M.~Dalal, and S.~Levine.
\newblock Awac: Accelerating online reinforcement learning with offline datasets.
\newblock {\em arXiv preprint arXiv:2006.09359}, 2020.

\bibitem{nair2018overcoming}
A.~Nair, B.~McGrew, M.~Andrychowicz, W.~Zaremba, and P.~Abbeel.
\newblock Overcoming exploration in reinforcement learning with demonstrations.
\newblock In {\em IEEE international conference on robotics and automation}, 2018.

\bibitem{curiosity}
D.~Pathak, P.~Agrawal, A.~A. Efros, and T.~Darrell.
\newblock Curiosity-driven exploration by self-supervised prediction.
\newblock In {\em International Conference on Machine Learning}, 2017.

\bibitem{Rajeswaran2017LearningCD}
A.~Rajeswaran, V.~Kumar, A.~Gupta, J.~Schulman, E.~Todorov, and S.~Levine.
\newblock Learning complex dexterous manipulation with deep reinforcement learning and demonstrations.
\newblock {\em Proceedings of Robotics: Science and Systems}, 2018.

\bibitem{bridge_offline_online}
P.~Rashidinejad, B.~Zhu, C.~Ma, J.~Jiao, and S.~Russell.
\newblock Bridging offline reinforcement learning and imitation learning: A tale of pessimism.
\newblock In {\em Advances in Neural Information Processing Systems}, 2021.

\bibitem{ppo}
J.~Schulman, F.~Wolski, P.~Dhariwal, A.~Radford, and O.~Klimov.
\newblock Proximal policy optimization algorithms.
\newblock {\em arXiv preprint arXiv:1707.06347}, 2017.

\bibitem{state_entropy_lili}
Y.~Seo, L.~Chen, J.~Shin, H.~Lee, P.~Abbeel, and K.~Lee.
\newblock State entropy maximization with random encoders for efficient exploration.
\newblock In {\em International Conference on Machine Learning}, 2021.

\bibitem{alphago}
D.~Silver, A.~Huang, C.~J. Maddison, A.~Guez, L.~Sifre, G.~Van Den~Driessche, J.~Schrittwieser, I.~Antonoglou, V.~Panneershelvam, M.~Lanctot, et~al.
\newblock Mastering the game of go with deep neural networks and tree search.
\newblock {\em nature}, 2016.

\bibitem{hyq}
Y.~Song, Y.~Zhou, A.~Sekhari, D.~Bagnell, A.~Krishnamurthy, and W.~Sun.
\newblock Hybrid {RL}: Using both offline and online data can make {RL} efficient.
\newblock In {\em The Eleventh International Conference on Learning Representations}, 2023.

\bibitem{jsrl}
I.~Uchendu, T.~Xiao, Y.~Lu, B.~Zhu, M.~Yan, J.~Simon, M.~Bennice, C.~Fu, C.~Ma, J.~Jiao, S.~Levine, and K.~Hausman.
\newblock Jump-start reinforcement learning.
\newblock In {\em Proceedings of the 40th International Conference on Machine Learning}, 2023.

\bibitem{vecerik2017leveraging}
M.~Vecerik, T.~Hester, J.~Scholz, F.~Wang, O.~Pietquin, B.~Piot, N.~Heess, T.~Roth{\"o}rl, T.~Lampe, and M.~Riedmiller.
\newblock Leveraging demonstrations for deep reinforcement learning on robotics problems with sparse rewards.
\newblock {\em arXiv preprint arXiv:1707.08817}, 2017.

\end{thebibliography}
\bibliographystyle{abbrv}


\end{document}